\pdfoutput=1
\documentclass[11pt]{article}

\usepackage[]{acl}
\usepackage{times}
\usepackage{latexsym}

\usepackage{helvet}
\usepackage{courier}
\usepackage{multirow}
\usepackage{enumitem}
\usepackage{hyperref}
\usepackage{url}
\usepackage{amsfonts}
\usepackage{dsfont}
\usepackage{booktabs}
\usepackage{caption}
\usepackage{subcaption}
\usepackage{adjustbox}
\usepackage{multicol}
\usepackage{amssymb}
\usepackage{graphicx}
\usepackage{mathtools}
\usepackage{amsmath}
\usepackage{tipa} % concatenation symbol
\usepackage{svg}
\usepackage{txfonts}
\usepackage{pifont}% http://ctan.org/pkg/pifont

\newcommand{\cmark}{\ding{51}}%
\newcommand{\xmark}{\ding{55}}%
\newcommand{\cross}{\ding{61}}%
\newcommand{\mb}[1]{\textcolor{red}{#1}}
\newcommand{\ch}[1]{\textcolor{blue}{#1}}

\usepackage{times}
\usepackage{latexsym}

\usepackage[T1]{fontenc}
\usepackage[utf8]{inputenc}

\usepackage{microtype}

\title{Regularization for Long 
Named Entity Recognition}

\author{Minbyul Jeong \\
  Korea University \\
  \texttt{minbyuljeong@korea.ac.kr} \\\And
  Jaewoo Kang \\
  Korea University \\
  \texttt{kangj@korea.ac.kr} \\}

\begin{document}
\maketitle
\begin{abstract}
When performing named entity recognition (NER), entity length is variable and dependent on a specific domain or dataset.
Pre-trained language models (PLMs) are used to solve NER tasks and tend to be biased toward dataset patterns such as length statistics, surface form, and skewed class distribution.
These biases hinder the generalization ability of PLMs, which is necessary to address many unseen mentions in real-world situations.
We propose a novel debiasing method RegLER to improve predictions for entities of varying lengths.
To close the gap between evaluation and real-world situations, we evaluated PLMs on partitioned benchmark datasets containing unseen mention sets.
Here, RegLER shows significant improvement over long-named entities that can predict through debiasing on conjunction or special characters within entities.
Furthermore, there is a severe class imbalance in most NER datasets, causing easy-negative examples to dominate during training, such as \textit{'The'}.
Our approach alleviates skewed class distribution by reducing the influence of easy-negative examples.
Extensive experiments on the biomedical and general domains demonstrated the generalization capabilities of our method.
To facilitate reproducibility and future work, we release our code.~\footnote{https://github.com/minstar/RegLER}
\end{abstract}

\section{Introduction}

Named entity recognition (NER) is a core task in information extraction, that is essential for many downstream tasks such as question answering \cite{molla2006named,lee-etal-2020-answering}, machine translation \cite{babych2003improving,ugawa2018neural}, entity linking \cite{martins-etal-2019-joint}, and event extraction \cite{wadden2019entity}.
In recent years, with the proliferation of pre-trained language models (PLMs) \cite{devlin2019bert,liu2019roberta,lee2020biobert}, the performance of NER datasets plateaued.

There are several ways to analyze a generalization behavior to see whether the PLMs are able to predict various unseen mentions, such as: breaking down a holistic performance into fine-grained categories \cite{fu-etal-2020-interpretable,fu2020rethinking}, splitting benchmark datasets based on seen and unseen mentions \cite{lin2020rigourous}, and partitioning unseen mentions into synonyms and concepts in the biomedical domain \cite{kim2021your}.
In addition, it has been proven that PLMs tend to focus on dataset biases to solve particular benchmark datasets~\cite{jia2017adversarial,gururangan2018annotation,nie2020adversarial}.

Exploiting dataset biases hinders the generalization capacity of PLMs to address unseen entities, which is critical in real-world situations.
Thus, current debiasing methods have been suggested to prevent the overfitting of superficial cues in training datasets \cite{clark2019don, kim2021your}.
For NER tasks, we found two problems to apply this directly on PLMs.
First, \citet{clark2019don} introduced a product-of-expert to scale down the gradients of biased examples, but they ignored a class prior because the class distributions of particular datasets is uniformly distributed.
In contrast, there is a severe class imbalance in the NER task, thus the dominating influence of easy-negative examples must be addressed~\cite{meng2019dsreg,li2020dice,mahabadi2020end}.
Second, in biomedical named entity recognition (BioNER), \citet{kim2021your} enhanced generalizability through the debiasing framework which replaces a bias-only model as a class distribution given word frequency.
However, it is impossible to explicitly provide a training signal to out-of-vocabulary (OOV) words that do not appear in the training.
For example, it is difficult to predict \textit{'Latent BHRF1 transcripts'} as an entity because the word \textit{'Latent'} does not occur in the training dataset but appears in the test dataset.

\begin{figure}[]
 \centering
 \includegraphics[width=180px]{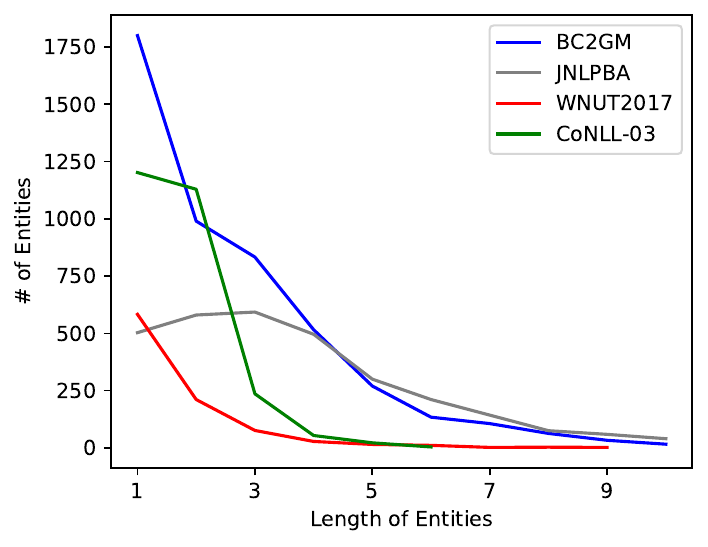}
%  \caption{Biomedical entities are longer compared to general domain entities.}
 \caption{Entity length is variable and dependent on a specific domain or dataset.}
 \label{fig:ablation length of entities}
\end{figure}

In this paper, we introduce a debiasing method RegLER that regularizes models to predict long-named entities by alleviating class imbalance and OOV issues.
When designing a class prior, we use pointwise mutual information (PMI) to measure the correlation score between a word and its class~\cite{gururangan2018annotation}.
Class imbalance is alleviated by reducing the influence of easy-negative examples such as \textit{'The'}.
We also leverage subword-level PMI scores to handle the dataset shift.
For instance, a word \textit{'Latent'} can be split into subwords of \textit{'Late'} and \textit{'\#\#nt'}, and the word \textit{'Late'} only occurs with an 'O' label which can cause it to not be predicted as an entity.
Thus, we debias the class distribution of the word \textit{'Late'} to predict it as an entity without the 'O' label.

\begin{figure*}[t]
 \centering
 \includegraphics[width=460px]{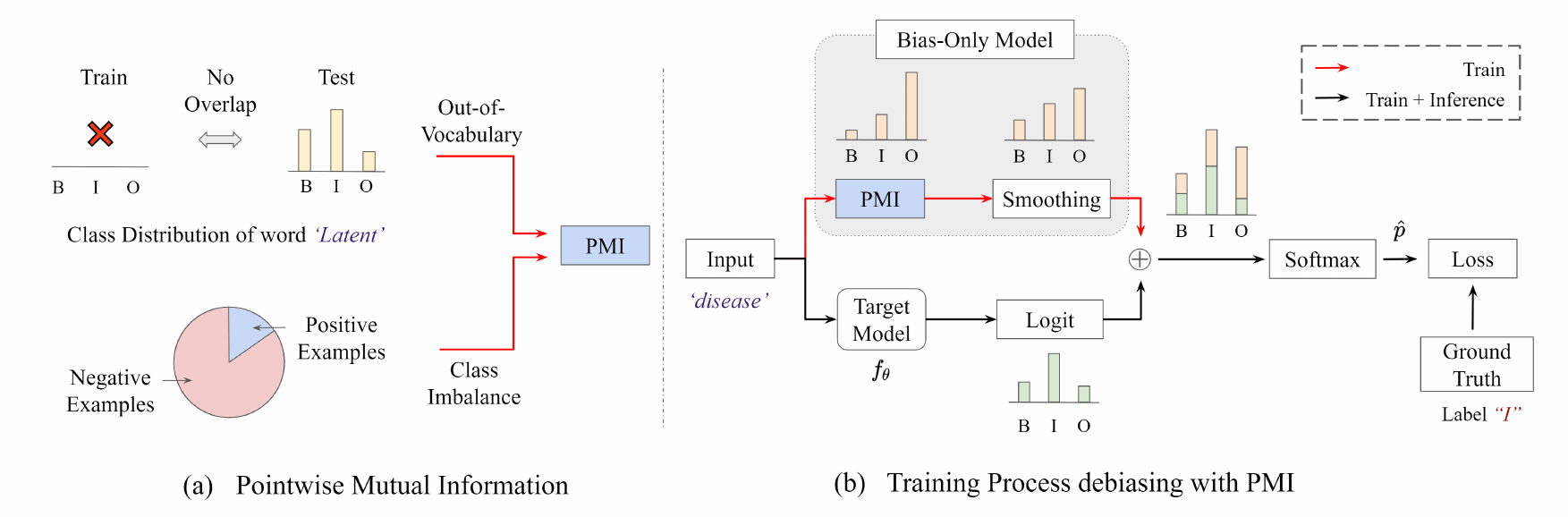}
 \caption{The overview of RegLER. (a) PMI reflects a solution of class imbalance, and the subword frequency solves an out-of-vocabulary (OOV) issue. For example, we depict the OOV issue with the word \textit{Latent} that does not appear in the training dataset but occurs in the test dataset. 
 We designed a class prior to reflecting the class imbalance inherent in the NER task. (b) We substitute a bias-only model with PMI statistic.
 For example, the word \textit{disease} is mainly classified with an \textit{I} (or inside) label, but the class distribution focuses on an \textit{O} (or outside) label.
 Our bias-only model reduces the importance of the label \textit{I} by adding the PMI score to the logit value of the target model ($f_\theta$). 
 Subsequently, we use temperature scaling \protect\cite{guo2017calibration} to smooth the class distribution of the normalized PMI depending on the entity length. The red line indicates that computing the PMI is only added during the training time.}
 \label{fig:model architecture}
\end{figure*}

Finally, previous studies have suggested that entity length can affect the predictions of the models~\cite{fu2020rethinking, hong2020dtranner, wei2020multichannel}.
% As depicted in Figure~\ref{fig:ablation length of entities}, we observe that biomedical entities are longer compared to the general domain entities because of adjectives or conjunctions within the entity~\cite{wei2020multichannel, cho2020combinatorial}.
As depicted in Figure~\ref{fig:ablation length of entities}, we observe that the entity length is variable and dependent on a specific domain or dataset.
Biomedical datasets contain longer entities than general domain datasets on average because of  adjectives or conjunctions found within the entity~\cite{wei2020multichannel, cho2020combinatorial}.
For example, an entity \textit{'foot - and - mouth - disease virus L. protease'} consists of 10 words because of the adjectives (foot-and-mouth) and conjunctions (- or and).
Thus, we formulate a target model (PLM in our case) that has flexibility depending on whether the entity length is short or long.
In other words, our RegLER method adaptively smooths the class distribution of the entity length.
The key intuition of our approach, depicted in Figure~\ref{fig:model architecture}, is to explicitly incorporate the dataset biases into the target model to increase robustness against these biases.

Our contributions are as follows: 
(1) We introduce a model-agnostic debiasing method, RegLER, to regularize the model to improve prediction for long-named entities by alleviating severe class imbalance and OOV issues in the NER task.
(2) We demonstrate that our method significantly improves the ability of synonym and concept generalization and shows competitive in-domain performance among the biomedical and general domains.
% (1) We introduce a model-agnostic debiasing method RegLER to regularize model to predict well for long-named entities. 
% (1) We propose a model-agnostic debiasing method RegLER that the statistics of PMI can replace a bias-only model. 
% (2) Our approach alleviates the class imbalance inherent for the task of NER and utilizes a subword frequency to tackle the OOV issue. 
% (3) Our method adaptively smooths a class distribution of the bias-only model and achieves higher improvement on long-named and complex structure entities.
% (4) Our experiments demonstrate that our method significantly improves the generalization ability and shows competitiveness in-domain performance in biomedical NER datasets.

\section{Backgrounds}
\paragraph{Task Formulation}
A NER is a task that identifies a word or phrase that corresponds to instances such as a person, location, organization, or miscellaneous.
NER mainly focuses on extracting and classifying named entities in a corpus with predefined entity tags.
We use the BIO tagging scheme \cite{ramshaw1999text}, where B and I labels are concatenated with all target entity types and the O label indicates that a token belongs to no chunk, for example, B-PER and B-LOC.
Formally, a sentence consists of an input sequence \textbf{X} $=$ \textbraceleft$x_1, x_2, \ldots, x_N$\textbraceright with length $N$, its ground truth labels of the input sequence \textbf{Y} $=$ \textbraceleft$y_1, y_2, \ldots, y_N$\textbraceright, and its predicted labels of the output sequence $\hat{\textbf{Y}}$ $=$ $\{\hat{y}_1, \hat{y}_2, \ldots, \hat{y}_N\}$ where $y_i$ $\in$ \textbraceleft$1, 2, \ldots, C$\textbraceright~ and $C$ is the number of classes. 
There are two widely used decoding strategies: tag-independent decoding (i.e., softmax) and tag-dependent decoding (i.e., conditional random field~\cite{lafferty2001conditional}).
The goal of NER is to predict mentions by assigning the output label $\hat{y}_{i}$ for each token $x_{i}$ using the following decoding strategies: $P(\hat{y}_i | \textbf{X})$ or $P(\hat{y}_i | \textbf{X}, \hat{y}_1, \hat{y}_2, \ldots, \hat{y}_{i-1})$.

\paragraph{Partitioning Benchmark Datasets}
\label{sec:partition}
Given the annotated entities in the training dataset, we created an entity dictionary to split the evaluation dataset.
In other words, using this dictionary generates two mention sets: mentions those that are seen and unseen during training~\cite{lin2020rigourous, kim2021your}.
In particular BioNER task, a raw dataset suggests the CUI of each entity that signifies the same concept (or meaning) of different entities, forming a concept dictionary.
We define our evaluation dataset to be split based on these dictionaries as follows,

\begin{itemize}[leftmargin=0.5in]
    \item $\textsc{Mem} := \{ M | M \in E_{tr}(M)\}$
    \item $\textsc{Unseen} := \{M|M \notin {E_{tr}}(M) \}$
    \item $\textsc{Syn} := \{ M | M \notin E_{tr}(M), M \in C_{tr}(M) \}$
    \item $\textsc{Con} := \{ M | M \notin E_{tr}(M), M \notin C_{tr}(M) \}$,
\end{itemize}
where $E_{tr}(M)$ and $C_{tr}(M)$ refer to the entity and concept dictionary of the training dataset respectively, and $M$ indicates the surface form of the entity.
Memorization (\textsc{Mem}) contains mentions that appear in the training dataset.
In contrast, \textsc{Unseen} contains mentions that do not occur in the training dataset.
More specifically, in the BioNER dataset, we can further partition the \textsc{Unseen} mention set into synonym generalization (\textsc{Syn}) and concept generalization (\textsc{Con})~\cite{kim2021your}.
We measure the generalization ability using these partitioned benchmark datasets.

\paragraph{Evaluation Metrics}
Evaluation of NER is divided into two parts: boundary detection and type identification \cite{li2020survey}.
General domain NER identifies and classifies text spans into human-annotated categories \cite{sang2003introduction}.
However, BioNER mostly focuses on computing an entity-level boundary detection.
The common evaluation metric of NER and BioNER is to compute the entity-level metric using the micro-averaged F1 score:
$F1 = \frac{2*P*R}{P+R}$, where $P$ and $R$ are Precision and Recall respectively.
Along with the common evaluation metric (F1 score) as an in-domain performance, we use the evaluation metric of \cite{kim2021your} to report the recall score of partitioned mention sets because the predicted entities are likely not to exist in the annotated entities.
% Thus, we report the recall score of the partitioned mention sets.

\paragraph{Debiasing Framework}
% biased model (ensemble) description + prior distribution statistics
% OOV problem, class imbalance, word occurrence
We illustrate the overall framework of our debiasing method RegLER in Figure~\ref{fig:model architecture}(b).
Given the input sequence of tokens \textbf{X} and its ground truth labels \textbf{Y}, a model has to predict the labels of the output sequence $\hat{\textbf{Y}}$.
As a simple ensemble method, we use a bias product that trains a product of experts \cite{hinton2002training,smith2005logarithmic} and adds the log probabilities of the bias-only model and target model $f_\theta$ \cite{clark2019don,he2019unlearn}.
It is intuitive that the target model can prevent a focus on dataset biases during training.
We compute a final probability as follows:

\begin{equation}
    \hat{p_{i}} = softmax(log(p_i) + log(b_i)),
\end{equation}
where $\hat{p_i} \propto p_i \cdot b_i$, a $log(p_i)$ is the log probability of the target model $f_\theta$, in which we desire to avoid over-trust in the surface form, and $log(b_i)$ is the log probability of the bias-only model.

% Besides, Bias Product can be extended to Learned-Mixin and Learned-Mixin+H~\cite{clark2019don}.
% The Learned-Mixin method allows the target model to determine how much to trust dataset bias given input sequence.
% On the contrary, the Learned-Mixin+H method used to prevent the target model from neglecting bias probability $log(b_i)$ by adding an entropy penalty.
% Detailed explanations are described in Appendix~\ref{app:learned-mixin}.
% Detailed explanations are described in Appendix.

% Overall, our method can operate on these ensemble frameworks without introducing additional parameters.
By converting a bias-only model to word level  statistics~\cite{ko2020look, kim2021your}, the model can operate on these ensemble frameworks without introducing additional parameters and we adopt this to address a various dataset patterns.
We define a loss function for our model as follows:

\begin{equation}
    L = -\frac{1}{N}\sum_{i=1}^{N} y_{i} \cdot \hat{p_i},
\end{equation}
\begin{equation}
    \hat{p_i} = softmax(log(p_i) + log(b_i)), 
\end{equation}
where $N$ is the length of an input sequence.
This loss function minimizes the negative log-likelihood of the distribution of final probability $\hat{p_i}$ and its ground-truth label $y_i$.
Note that $log(b_i)$ is only added during the training process, whereas it does not apply at inference time.
% This process is depicted with a red arrow in Figure~\ref{fig:model architecture}.

\section{Motivation}

\paragraph{Long and Complex Structure Entities}
\label{sec:longnamed}
In terms of entity length, there are several error cases in which longer terms are not well predicted.
This can be easily observed through Figure~\ref{fig:ablation length of entities}.
\citet{hong2020dtranner} shows that conjunctions or the special character "/" within an entity makes it challenging for the models to predict types of diseases and chemicals.
\citet{wei2020multichannel} demonstrated three primary reasons for error as follows: complex and long-named entities, annotation inconsistency on corpora, and abbreviations.
As the bias-only model does not over-trust the dataset bias, we hypothesize that relative smoothing on the bias-only model predictions can increase robustness when dealing with long entities.

\paragraph{Subword Frequency}
\label{sec:subword}
We depict the OOV issue in Figure~\ref{fig:model architecture}(a).
It is impossible to explicitly provide a training signal to OOV words that do not occur during training~\cite{leevy2018survey, li2020dice}.
% If a class distribution of a particular word is different in training and test datasets then models fail to reflect informative statistics~\cite{leevy2018survey,li2020dice}.
For example, the word \textit{'Latent'} does not appear in the training dataset but occurs as a part of an entity in the test dataset as \textit{'Latent BHRF1 transcripts'}. 
Thus, it is difficult to apply a meaningful statistic based on a word unit.
However, if subword statistics are applied, \textit{'Latent'} can be split into the subwords \textit{'Late'} and \textit{'\#\#nt'}.
As the word \textit{'Late'} occurs with only an 'O' label in the training dataset, debiasing to predict without using the 'O' label enables capturing the entity \textit{'Latent BHRF1 transcripts'}.
This example is fully detailed in Section~\ref{sec:informative}.

\paragraph{Modeling a Class Prior}
\label{sec:classprior}
A joint model of $f_\theta(\hat{\textbf{Y}}|\textbf{X})$ and $\frac{P(\textbf{Y}|\textbf{X})}{P(\textbf{Y})}$ is used by current debiasing methods~\cite{clark2019don}.
However, these methods use a bias-only model with learned parameters and ignore the $P(\textbf{Y})$ factor while designing because it does not need to reflect a class prior in a situation where the class distribution of a particular dataset (e.g., natural language inference) is uniformly distributed.
In contrast, class imbalance is severe in NER tasks, increasing the need to address the dominating influence of easy-negative examples~\cite{meng2019dsreg,li2020dice,mahabadi2020end}.
In Table~\ref{tab:all dataset statistics}, we can easily see that the ratio of negative to positive examples is extremely high in the NER tasks.
Thus, a class prior needs to be considered to reduce the influence of such easy-negative examples.
% We report to Table~\ref{tab:all dataset statistics} separately due to the space issue.

\begin{table}[t]
\centering
\begin{adjustbox}{max width=0.48\textwidth}
\begin{tabular}{ l l c c c c }
\toprule
\multicolumn{1}{l}{Dataset} & \multicolumn{1}{l}{Entity Type} & \multicolumn{1}{c}{\begin{tabular}[c]{@{}c@{}}\# of\\ Sentences\end{tabular}} & \begin{tabular}[c]{@{}c@{}}Positive\\ Examples\end{tabular} & \begin{tabular}[c]{@{}c@{}}Negative\\ Examples\end{tabular} & Ratio \\
\midrule 
NCBI-disease & Disease & 8,230 & 13K & 203K & 15.47 \\ 
BC5CDR-disease & Disease & 12,058 & 14K & 313K & 22.29 \\
BC5CDR-chem & Drug/Chem. & 12,126 & 14K & 313K & 22.08 \\
BC4CHEMD & Drug/Chem. & 89,664 & 129K & 2.3M & 17.83 \\
BC2GM & Gene/Protein & 21,370 & 44K & 535K & 11.96 \\ 
JNLPBA & Gene/Protein & 27,051 & 122K & 637K & 5.20 \\
LINNAEUS & Species & 21,459 & 4.3K & 478K & 110.71 \\
Species-800 & Species & 8,723 & 6.7K & 220K & 32.79 \\
CoNLL-2003 & General & 14,986 & 34K & 170K & 4.98 \\
WNUT2017 & General & 3,394 & 15K & 74K & 4.79 \\
\bottomrule
\end{tabular}
\end{adjustbox}
\caption{Data statistics of the NER tasks. All datasets are listed with the details: entity type, number of sentences, positive examples (B and I labels), negative examples (O labels), and the ratio of negative examples to positive examples (negative examples/positive examples).}
\label{tab:all dataset statistics}
\end{table}

\section{Methods}
We start by introducing our debiasing method RegLER, which utilizes PMI as a bias-only model to alleviate the OOV issue and class imbalance.
For long entities, we provide an explanation for smoothing the class distribution via temperature scaling~\cite{guo2017calibration}.

\subsection{Pointwise Mutual Information}
\label{sec:pmi}
To cope with the class-distribution shift of words between the training and test datasets\footnote{PMI was used as a correlation score between the word and class in a particular dataset \cite{gururangan2018annotation}
}, we compute the PMI between the subword and class in the training dataset as follows:

\vspace{-0.2cm}
\begin{equation}
\hspace*{-0.2cm}
PMI(subword, class) = log\frac{P(subword,class)}{P(subword) \cdot P(class)}.
\label{eq:PMI}
\end{equation}
For a numerical issue, we apply add-K smoothing to each subword frequency\footnote{K is a hyperparameter and we use K=100, similar to \cite{gururangan2018annotation} to emphasize correlations of word and class.}.

Overall, PMI captures the correlation between subwords and their corresponding classes.
We further suggest a method to understand how each mathematical expression operates.
A joint distribution $P(subword, class)$ assigns low training signals to subwords in which class distributions are dominant for a particular label.
However, $P(subword, class)$ focuses on frequent subwords independent of class, such as \textit{'disease'} or \textit{'deficiency'}. 
To handle this problem, a $P(subword)$ alleviates the situation in which subwords appear frequently regardless of their class. 
Finally, $P(class)$ enhances regularization when the class imbalance between positive examples (B and I labels) and negative examples (O labels) is severe.
Because there are many easy-negative examples, we want to prevent the training signals from being ruled by these easy-negatives.

As the PMI is much larger than the logit value of the target model $f_\theta$, we normalize the PMI to a probability distribution. 
We can then compute the normalized PMI (i.e., $PMI_{ic}$) and the final probability $\hat{p_i}$ as follows:

\begin{equation}
    PMI_{ic} = \frac{exp(PMI_{ic})}{\sum_{c'=1}^{C}exp(PMI_{ic'})}
\end{equation}
\begin{equation}
    \hat{p_{i}} = softmax(log(p_i) + log(PMI_i))
\end{equation}
where $C$ is the number of classes in a particular dataset, and $PMI_{ic}$ indicates the probability of the PMI corresponding to each class $c$.
$PMI_i = \{ PMI_{i1}, PMI_{i2} \ldots, PMI_{iC} \}$ is the class distribution of the PMI for each token $x_i$ in the training dataset.

\begin{table*}[t]
\begin{adjustbox}{width=\textwidth}
\begin{tabular}{ l c c c c c c c | c c c c | c c c }
\toprule
\multicolumn{1}{c}{\multirow{3}{*}{Method}} & \multicolumn{1}{c}{\multirow{3}{*}{\textsc{Subword}~}} &
\multicolumn{1}{c}{\multirow{3}{*}{\textsc{Class}~}} &
\multicolumn{1}{c}{\multirow{3}{*}{\textsc{Temp}~}} & \multicolumn{4}{c}{NCBI-disease} & \multicolumn{4}{c}{BC5CDR-chem} & \multicolumn{3}{c}{LINNAEUS} \\ \cmidrule{5-15} 
\multicolumn{1}{c}{} & \multicolumn{1}{c}{} & & \multicolumn{1}{c}{} & \begin{tabular}[c]{@{}c@{}}\textsc{Mem}\\ (R)\end{tabular} & 
\begin{tabular}[c]{@{}c@{}}\textsc{Syn}\\ (R)\end{tabular} & \begin{tabular}[c]{@{}c@{}}\textsc{Con}\\ (R)\end{tabular} & \begin{tabular}[c]{@{}c@{}}Total\\ (F1) \end{tabular} & \begin{tabular}[c]{@{}c@{}}\textsc{Mem}\\ (R)\end{tabular} & 
\begin{tabular}[c]{@{}c@{}}\textsc{Syn}\\ (R)\end{tabular} & \begin{tabular}[c]{@{}c@{}}\textsc{Con}\\ (R)\end{tabular} & \begin{tabular}[c]{@{}c@{}}Total\\ (F1) \end{tabular} & \begin{tabular}[c]{@{}c@{}}\textsc{Mem}\\ (R)\end{tabular} & 
\begin{tabular}[c]{@{}c@{}}\textsc{Unseen}\\ (R)\end{tabular} & \begin{tabular}[c]{@{}c@{}}Total\\ (F1) \end{tabular}  \\ \midrule
$SciFive-base$ \cross & \xmark & \xmark & \xmark & - & - & - & 89.4 & - & - & - & 94.2 & - & - & - \\
$Spark NLP$ \cross & \xmark & \xmark & \xmark & - & - & - & 89.1 & - & - & - & - & - & - & 86.3  \\
$KeBioLM$ \cross & \xmark & \xmark & \xmark & - & - & - & 89.1 & - & - & - & 93.3 & - & - & -  \\
$CL-KL$ \cross & \xmark & \xmark & \xmark & - & - & - & 89.0 & - & - & - & - & - & - & -  \\
$BioFLAIR$ \cross & \xmark & \xmark & \xmark & - & - & - & 88.9 & - & - & - & 93.5 & - & - & 87.0 \\
\midrule
$BioBERT$  & \xmark & \xmark & \xmark & 95.5 & 80.8 & 83.5 & 88.3 &  97.2 & 82.9 & 86.9 & 92.1 & 98.2 & 58.3 & 87.8 \\
$BioBERT + {BP}$ & \xmark & \xmark & \xmark & 93.9 & 81.1 & 84.1 & 88.1 & 96.8 & 83.1 & 88.0 & 92.0 & 98.1 & 60.4 & 87.0 \\ 
$BioBERT$ + RegLER & \cmark & \cmark & \cmark & \textbf{95.7} & \textbf{83.6} & 86.5 & 89.0 & 97.1 & 83.8 & 88.2 & 92.3 & 98.4 & \textbf{62.3} & 88.1 \\ \midrule
$SciBERT$ & \xmark & \xmark & \xmark & 94.9 & 82.1 & 80.0 & 88.4 & 97.2 & 76.5 & 83.5 & 91.4 & 96.3 & 50.7 & 86.8 \\
$SciBERT + {BP}$ & \xmark & \xmark & \xmark & 94.4 & 82.8 & 80.7 & 87.2 & 97.2 & 76.4 & 84.4 & 91.5 & 98.7 & 49.5 & 86.5 \\
$SciBERT$ + RegLER & \cmark & \cmark & \cmark & 94.5 & 83.3 & 80.5 & 88.1 & 97.1 & 78.0 & 84.4 & 91.8 & \textbf{99.0} & 55.5 & 87.7 \\ \midrule
$PubmedBERT$ & \xmark & \xmark & \xmark & 94.0 & 80.7 & 80.0 & 87.8 & 98.2 & 85.2 & 86.0 & 93.3 & 98.2 & 55.9 & 87.5 \\
$PubmedBERT + {BP}$ & \xmark & \xmark & \xmark & 94.1 & 81.8 & 81.6 & 87.4 & 98.0 & 87.1 & 86.8 & 93.3 & 98.6 & 57.2 & 87.4 \\
$PubmedBERT$ + RegLER & \cmark & \cmark & \cmark& 95.3 & 82.6 & 83.4 & 88.3 & 98.5 & 85.8 & 87.1 & 93.8 & 98.9 & 59.0 & \textbf{88.5} \\ \midrule
$BioLM-base$ & \xmark & \xmark & \xmark & 93.5 & 82.4 & 86.2 & 88.0 & 98.6 & 89.6 & 88.6 & 92.8 & 98.3 & 55.9 & 88.0 \\
$BioLM-base + {BP}$ & \xmark & \xmark & \xmark & 93.5 & 82.5 & 86.2 & 87.9 & 97.8 & 89.9 & 88.6 & 92.9 & 97.5 & 56.1 & 86.1 \\
$BioLM-base$ + RegLER & \cmark & \cmark & \cmark & 93.5 & 83.1 & 86.7 &  88.6 & \textbf{98.7} & 90.0 & 88.9 & 93.2 & 97.9 & 56.2 & 87.4 \\
\midrule
$BioLM-large$ & \xmark & \xmark & \xmark & 94.8 & 80.4 & 83.7 & 88.5 & 98.3 & 88.4 & 89.0 & 94.1 & 98.8 & 55.2 & 87.5 \\
$BioLM-large + {BP}$ & \xmark & \xmark & \xmark & 93.5 & 81.2 & 85.0 & 88.5 & 98.4 & 89.5 & 88.9 & 94.2 & 98.9 & 57.5 & 86.3 \\
$BioLM-large$ + RegLER & \cmark & \cmark & \cmark & 95.1 & 83.2 & \textbf{87.4} & \textbf{89.5} & \textbf{98.7} & \textbf{90.3} & \textbf{89.1} & \textbf{94.4} & \textbf{99.0} & 58.2 & 87.5 \\
\bottomrule
\end{tabular}
\end{adjustbox}
\caption{Performance of the debiasing method on the biomedical domain NER datasets. Each dataset is partitioned into memorization (\textsc{Mem}), synonym generalization (\textsc{Syn}), and concept generalization (\textsc{Con}). 
We use \cmark and \xmark ~to show the components that are used. \cross ~signifies the reported performance on the manuscript. Best performances are shown in bold.}
\label{tab:maintable 1}
\end{table*}

\subsection{Bias-weighted Scaling on Long-Named and Complex Entities}
\label{sec:scaling}
As illustrated in Figure~\ref{fig:model architecture}, our RegLER method involves adaptively smoothing the class distribution of PMI using temperature scaling~\cite{guo2017calibration}.
Biomedical entities have long names and complex structures that contain conjunctions or adjectives~\cite{wei2020multichannel,cho2020combinatorial}. 
Thus, we define a temperature $T_i$ that depends on the length $L_i$ of the entity to smooth the class distribution of the PMI.
Given the PMI probability $PMI_{ic}$, the new statistic is defined as below,

\begin{equation}
    T_i := \begin{cases}
        1 + \lambda \cdot L_i, & \text{if $L_i > 1$} \\
        1, & \text{otherwise}
    \end{cases}
\end{equation}
\begin{equation}
    PMI_{ic} = \frac{exp(PMI_{ic} / T_i)}{\sum_{c'=1}^{C}exp(PMI_{ic'} / T_i)}.
\end{equation}
$T_i$ is the temperature~\cite{guo2017calibration} of the token $x_i$ to smooth the class distribution of $PMI_{ic}$ depending on the entity length $L_i$, and $\lambda$ is a hyperparamter.
As $T \text{→} ~\infty$, the probability $PMI_{ic}$ approaches $1 / C$, which represents the maximum uncertainty.
Conversely, as $T \text{→} ~ 0$, the probability collapses to a point mass.

Notably, as PMI requires task-related information, it does not introduce additional memory requirements.
We can also choose other statistics, such as the random value or word frequency corresponding to its class.
However, we observe that using PMI achieves the best results.
Moreover, improving the generalizability of our method appears to be beneficial for other class-imbalance tasks.

\section{Experiments}
In this section, we evaluate our debiasing method RegLER on the biomedical and general domain datasets of the NER task.
We address three main questions in the experiments as follows: 
(1) How effective is our method as a bias-only model using length-dependent scaling, subword frequency, and the class prior? (2) Does bias-weighted scaling predict entities with long names and complex structures? (3) What components can support the capacity of generalization?

\subsection{Experimental Settings}

\paragraph{Dataset}
We evaluated our debiasing method on ten benchmark NER datasets including eight biomedical domains and two general domains.
For the biomedical domains, we used eight datasets: disease entities NCBI-disease \cite{dougan2014ncbi}, BC5CDR-disease \cite{li2016biocreative}, drug and chemical entities BC5CDR-chemical \cite{li2016biocreative}, BC4CHEMD \cite{krallinger2015chemdner}, gene and protein entities BC2GM \cite{smith2008overview}, JNLPBA \cite{kim2004introduction}, and species entities LINNAEUS \cite{gerner2010linnaeus}, and Species-800 \cite{pafilis2013species}. 
We followed the pre-processed version of the previous work \cite{crichton2017neural}.\footnote{https://github.com/cambridgeltl/MTL-Bioinformatics-2016}
For the general domain, we used two English datasets: the CoNLL-2003 shared benchmark dataset \cite{sang2003introduction} and WNUT-2017 which focuses on identifying unusual, previously unseen entities \cite{derczynski2017results}.
We focused on improving both the in-domain performance and recall of our partitioned datasets (\textsc{Mem}, \textsc{Syn}, \textsc{Con}, and \textsc{Unseen}).
Descriptions of data statistics are noted in Appendix~\ref{app:dataset statistic}.
% Descriptions of data statistics are suggested in Appendix.

\paragraph{Baselines}
Our target models are composed of BERT and RoBERTa architectures, which have been widely used in many studies on NER.
We used BioBERT~\cite{lee2020biobert}, SciBERT~\cite{beltagy2019scibert}, PubmedBERT~\cite{liu2020domain}, and BioLM~\cite{lewis2020pretrained} as our target models. 
% We leverage BERT token embeddings to train Bi-LSTM~\cite{huang2015bidirectional} architecture with CRF decoder ($BiLSTMCRF$). 
To compare our experimental results, we report various strong models with written scores.
These models are enumerated as follows: SciFive-base~\cite{phan2021scifive}, Spark NLP~\cite{kocaman2021spark}, KeBioLM~\cite{yuan2021improving}, CL-KL~\cite{wang2021improving}, and  BioFLAIR~\cite{sharma2019bioflair}.
In the bias-only model, we compare RegLE with a class distribution of word frequency~\cite{kim2021your} which is suggested as row 2 of each block in Table~\ref{tab:maintable 1}.
% To demonstrate a regularize effect of task-related statistics (e.g., \textsc{Word} and PMI), we add the \textsc{Rand} as a baseline of the bias-only model. 
Rather than attempting PMI solely on the Bias-Product ($BP$), we extend our experiments to show how our method RegLE can operate with other debiasing methods such as Learned-Mixin ($LM$) and Learned-Mixin+H~\cite{clark2019don}. 
Detailed explanations are given in Appendix~\ref{app:learned-mixin}.
We denote \cmark and \xmark ~when using subword frequency (\textsc{Subword}), class prior (\textsc{Class}), and bias-weighted scaling (\textsc{Temp}).
Description of the training details and hyperparameters are provided in Appendix~\ref{app:hyperparameter}.
% Description of training details and hyperparameters are suggested in Appendix.

\begin{table*}[t]
\centering
\begin{adjustbox}{width=.9\textwidth}
\begin{tabular}{ l c c c c c c c | c c c c }
\toprule
\multicolumn{1}{c}{\multirow{3}{*}{Method}} & \multicolumn{1}{c}{\multirow{3}{*}{\textsc{Subword}~}} & \multicolumn{1}{c}{\multirow{3}{*}{\textsc{Class}~}} &
\multicolumn{1}{c}{\multirow{3}{*}{\textsc{Temp}~}} & \multicolumn{4}{c}{NCBI-disease} & \multicolumn{4}{c}{BC5CDR-chem} \\ \cmidrule{5-12} 
\multicolumn{1}{c}{} & \multicolumn{1}{c}{} & \multicolumn{1}{c}{} & \multicolumn{1}{c}{} & \begin{tabular}[c]{@{}c@{}}\textsc{Mem}\\ (R)\end{tabular} & 
\begin{tabular}[c]{@{}c@{}}\textsc{Syn}\\ (R)\end{tabular} & \begin{tabular}[c]{@{}c@{}}\textsc{Con}\\ (R)\end{tabular} & \begin{tabular}[c]{@{}c@{}}Total\\ (F1) \end{tabular} & \begin{tabular}[c]{@{}c@{}}\textsc{Mem}\\ (R)\end{tabular} & 
\begin{tabular}[c]{@{}c@{}}\textsc{Syn}\\ (R)\end{tabular} & \begin{tabular}[c]{@{}c@{}}\textsc{Con}\\ (R)\end{tabular} & \begin{tabular}[c]{@{}c@{}}Total\\ (F1) \end{tabular} \\ \midrule
$BioLM-large$ + RegLER & \xmark & \xmark & \xmark & 93.5 & 81.2 & 85.0 &  88.1 & 98.4 & 89.5 & 88.9 & 94.2 \\ 
$BioLM-large$ + RegLER & \cmark & \xmark & \xmark & 94.2 & 80.8 & 84.6 & 88.6 & \textbf{98.7} & 87.4 & 86.7 & 93.2 \\ 
$BioLM-large$ + RegLER & \xmark & \cmark & \xmark & 94.7 & 82.4 & 84.8 & 88.6 & 98.6 & 88.5 & 87.5 & 93.5 \\ 
$BioLM-large$ + RegLER & \xmark & \xmark & \cmark & 94.0 & 81.3 & 84.5 & 88.7 & 97.7 & 85.4 & 86.8 & 93.8 \\ 
$BioLM-large$ + RegLER & \cmark & \cmark & \xmark & \textbf{95.2} & 82.5 & \textbf{88.1} & 88.5 & 98.6 & 90.1 & \textbf{89.2} & 93.6 \\ 
$BioLM-large$ + RegLER & \cmark & \xmark & \cmark & 95.0 & 82.4 & 84.3 & 88.8 & 98.6 & 88.3 & 87.4 & 94.1 \\ 
$BioLM-large$ + RegLER & \xmark & \cmark & \cmark & 94.8 & \textbf{83.5} & 86.2 & 89.1 & 98.4 & 89.9 & 88.3 &  94.3 \\
$BioLM-large$ + RegLER & \cmark & \cmark & \cmark & 95.1 & 83.2 & 87.4 & \textbf{89.5} & \textbf{98.7} & \textbf{90.3} & 89.1 & \textbf{94.4} \\
\bottomrule
\end{tabular}
\end{adjustbox}
\caption{Ablation study of three components on biomedical domain: subword frequency~(\textsc{Subword}), class prior~(\textsc{Class}), and length dependent scaling~(\textsc{Temp}). Best performances are shown in bold.}
\label{tab:ablation table 1}
\end{table*}

\subsection{Effectiveness of Our Debiasing Method}
Experimental results on the biomedical domain are summarized in Table~\ref{tab:maintable 1}.
% Table~\ref{tab:maintable 1} and \ref{tab:maintable 2} show our results on both biomedical and general domain.
Using three components, namely \textsc{Subword}, \textsc{Class}, and \textsc{Temp}, showed significant improvements in recall on \textsc{Syn} and \textsc{Con}, as well as overall improvements to in-domain performance (F1).
In the first block, we report five experimental results to compare our debiasing method RegLER.
From the second to the sixth block, we probed the performance of various language models trained on the Bias-Product ($BP$) framework.

Compared to the fine-tuning language model (row 1 in each block), the absolute improvements of our method RegLER in \textsc{Syn}, \textsc{Con}, and in-domain performance (F1) were approximately 1.34\%, 1.4\% and 0.47\%, respectively.
In the $BP$ framework, the absolute improvements of our method in \textsc{Syn}, \textsc{Con}, and F1 were approximately 1.01\%, 0.82\% and 0.66\% compared to the \cite{kim2021your} method, respectively.

While the $BP$ framework is vulnerable to preserving memorization ability and in-domain performance, our method achieves competitive performance on all datasets and achieves state-of-the-art performance on three datasets.
Because our method RegLER targets capturing long-named entities, we achieve the ability to generalize but fall behind various baselines in the general domain.
We thus report our performance for the general domain in Appendix~\ref{app:general domain}.
We conduct five runs with different seeds for all experiments.
The t-tests indicate that p $<$ 0.05.
We report on other datasets owing to the space limit in Appendix~\ref{app:experimental results}.

\subsection{Quantitative Analysis}
Overall, adopting three components not only boosts the performance of memorization ability in all cases but also improves  generalizability by preventing dataset bias.
In Table~\ref{tab:ablation table 1}, we note that the class prior (\textsc{Class}) and subword frequency (\textsc{Subword}) improve the memorization ability (\textsc{Mem}) but decrease the capacity of concept generalization (\textsc{Con}).
However, a combination of \textsc{Subword} and \textsc{Class} shows a consistent improvement in \textsc{Syn} and \textsc{Con}.
Furthermore, applying temperature scaling (\textsc{Temp}) based on the entity length enhances the ability of \textsc{Mem}. 
As we suggest the \textsc{Temp} to improve prediction for the long-named entities, there is an increase in in-domain performance.
Although the performance improvement itself appears insignificant, we will show that the improvement is large for long entities in the next paragraph.
In summary, it seems that the three components maintain an equilibrium to increase the robustness of the model, allowing for a high capacity for generalization and memorization.

\begin{figure}[!t]
 \centering
 \includegraphics[width=220px]{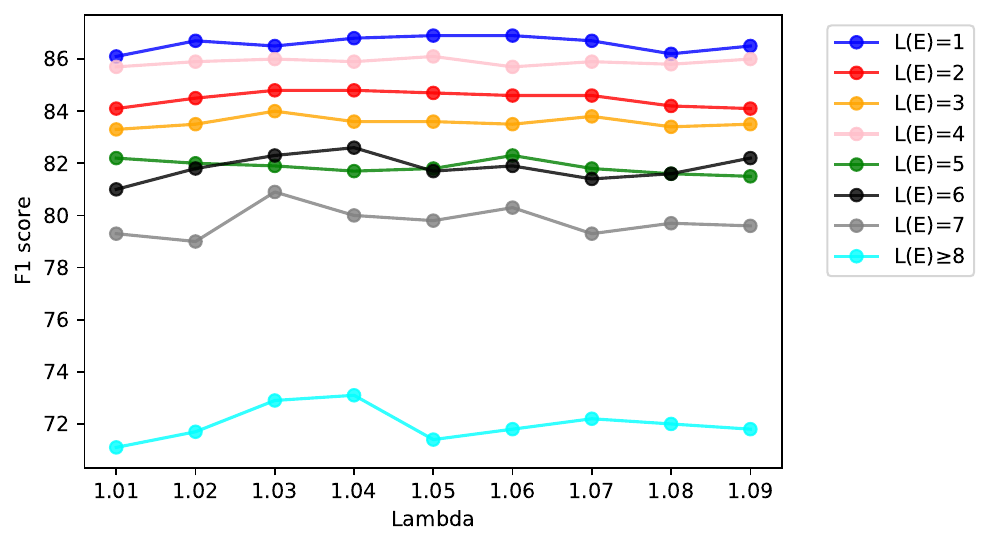}
 \caption{Bias-weighted scaling on long-named and complex entities. L(E) indicates an entity length.}
 \label{fig:length ablation}
\end{figure}

\begin{table*}[t]
\centering
\begin{adjustbox}{width = \textwidth}
\begin{tabular}{l l l l l l l l l l l l }
\toprule
\multicolumn{12}{ c }{\textbf{JNLPBA (out-of-vocabulary 'Latent')}} \\ \midrule
\multicolumn{12}{ l }{\begin{tabular}[c]{@{}l@{}}
Sentence: \textbf{Latent BHRF1 transcripts} encoding \textbf{bcl - 2 homologue} and \textbf{BCRF1 transcripts} encoding \textbf{viral interleukin ( vIL )} \\ 
~~~~~~~~~~~~~~~~~ \textbf{- 10} were detected in one and two of eight patients , respectively .\\
$BioLM-large$: \mb{BHRF1 transcripts}, bcl - 2, BCRF1 transcripts, viral interleukin ( vIL ) - 10 \\
$BioLM-large$ + RegLER (+ None): \mb{BHRF1 transcripts}, bcl - 2 homologue, BCRF1 transcripts, viral interleukin ( vIL ) - 10 \\
$BioLM-large$ + RegLER (+ \textsc{Subword}): \ch{Latent BHRF1 transcripts}, bcl - 2 homologue, BCRF1 transcripts, viral interleukin ( vIL ) - 10\end{tabular}} \\
\midrule
\multicolumn{12}{ c }{\textbf{BC2GM (Long-Named Entity)}} \\ \midrule
\multicolumn{12}{ l }{\begin{tabular}[c]{@{}l@{}}
Sentence: \textbf{Rhinovirus 2A protease} and \textbf{foot - and - mouth - disease virus L protease} were used to analyze the association of \\ 
~~~~~~~~~~~~~~~~ \textbf{eIF4G} with \textbf{eIF4A} , \textbf{eIF4E} , and \textbf{eIF3} . \\ 
$BioLM-large$: Rhinovirus 2A protease, \mb{mouth - disease virus L protease}, eIF4G, eIF4A, eIF4E, eIF3 \\
$BioLM-large$ + RegLER (+ None): Rhinovirus 2A protease, \mb{mouth disease virus L protease}, eIF4G, eIF4A, eIF4E, eIF3 \\
$BioLM-large$ + RegLER (+ \textsc{Temp}): Rhinovirus 2A protease, \ch{foot - and - mouth - disease virus L protease}, eIF4G, eIF4A, eIF4E, eIF3 \end{tabular}} \\
\bottomrule
\end{tabular}
\end{adjustbox}
\caption{Example of dealing with the OOV issue and prediction for a long-named entity. Ground truth labels are in bold, incorrect answers are in red, and correct answers are in blue.}
\label{tab:qualitative example}
\end{table*}

\paragraph{Adjusting $\lambda$ for Temperature Scaling}
\label{exp:adjust lambda}
As shown in Figure~\ref{fig:length ablation}, for the BC2GM~\cite{smith2008overview} dataset, we conducted an ablation study by converting only the $\lambda$ during training.
Following previous work~\cite{fu-etal-2020-interpretable}, we adopt a fine-grained evaluation metric to partition the mention set based on the entity length.
In Figure~\ref{fig:ablation length of entities}, since biomedical entities are much longer than general entities, we suggest the plot to select entity length in the range from one to eight.
As observed, there is little change in the performance of the mention sets with a small length of entities.
However, in the long-named entities (E(L) $\geqq$ 8), we investigate that the F1 score improves by up to 2\%.
Detailed numerical values are reported in Appendix~\ref{app:lambda results}.

\subsection{Qualitative Analysis}
\label{sec:informative}
In this section, we present two intuitive examples to better understand our approach.
In Table~\ref{tab:qualitative example}, we demonstrate two examples of dealing with the OOV issue and predicting a long-named entity.
For example, the word \textit{'Latent'} does not appear in the training dataset but occurs in the evaluation dataset.
Thus, we cannot apply the class distribution of the word frequency based on the word unit.
However, \textit{'Latent'} can be split into \textit{'Late'} and \textit{'\#\#nt'} using a subword tokenizer.
Considering the word \textit{'Late'} highly co-occurs with the 'O' label, the bias-only model forces the target model to predict without taking the 'O' label into account.
This enables recognition of the entity \textit{'Latent BHRF1 transcripts'}.
By adopting bias-weighted scaling, our RegLER approach can capture entities that are long or exhibit complex structure.
This also allows for the prediction of \textit{foot - and - mouth - disease virus L protease} which includes conjunction (such as "and") and special symbol (such as "-").
In BioNER tasks, there are many more long entities, but we only show a relatively simple case to provide an intuition.

\section{Related work}
\paragraph{Named Entity Recognition} Named Entity Recognition (NER) has long been studied and received attention for being a core task of information extraction.
Initially, research of NER focused on constructing a model using handcrafted features \cite{zhou2002named,bender2003maximum,settles2004biomedical}.
Because this is labor-intensive, deep learning approaches automatically extract features and label a sequence \cite{dos2015boosting,lample2016neural,habibi2017deep}.
Recently, with the advent of pre-training on a large corpus, contextualized language models performed significantly better on benchmark datasets and reached a plateau in performance \cite{devlin2019bert,liu2019roberta,beltagy2019scibert,lee2020biobert}.

However, questions were raised about the generalization capabilities of pre-trained language models.
Specifically, in NER tasks, \citet{fu2020rethinking} divided the set of test entities into different subsets and broke the overall performance down into interpretable categories. 
To explore the generalization ability, \citet{lin2020rigourous} partitioned the benchmark datasets into seen and unseen mention sets to create situations of erasing name regularity, mention coverage, and context diversity.
In BioNER, \citet{kim2021your} split unseen mention sets into synonyms and concepts based on CUI overlap between the training and test datasets. 
The authors adopted a debiasing method to enhance generalization behavior.

\paragraph{Dataset Bias and Debiasing Methods}
% surface form에 치중, debiasing method 제안 => 기존 방법들과의 차이점 언급 및 장점 확실하게 어필
In NLP tasks, dataset bias is a superficial cue or surface form of text associated with specific labels.
Recently, pre-trained language models~\cite{devlin2019bert,liu2019roberta,lee2020biobert} have tended to exploit dataset biases to predict rather than understand the relationship between words and labels~\cite{gururangan2018annotation,mccoy2019right,geirhos2020shortcut,bender-koller-2020-climbing,utama2020towards,du2021interpreting}. 
This eventually led the advent of debiasing methods to enhance the generalization ability.
\citet{clark2019don} introduced a product-of-expert as a model-agnostic approach of debiasing method to scale down the gradients on the biased examples.
\citet{mahabadi2020end} proposed a bias-only model to focus on learning hard examples by reducing the relative importance of biased examples using focal loss.
\citet{schuster-etal-2019-towards} suggested a regularization method to lower the weights of the give-away phrases causing the bias.
As previous debiasing methods are vulnerable to preserving in-domain performance while improving out-of-domain performance, a confidence regularization~\cite{utama2020mind} method addresses this trade-off by connecting robustness against dataset biases and overconfident problems in neural models~\cite{papernot2016distillation,feng2018pathologies}.
Our work is motivated by a debiasing framework with a prior distribution of task-related statistics~\cite{ko2020look,kim2021your}, and we introduce another statistic that fits well for the task of class imbalance.

\section{Conclusion}
% class imbalance가 있는 다른 task에 적용이 가능해보인다 (opinion mining)
% In this paper, we introduce a debiasing method RegLER based on task-related subword statistics to alleviate the \textit{Out-Of-Vocabulary} issue for class imbalance tasks such as NER. 
% To this end, we adopt PMI by preventing solely focusing on dataset biases.
% We argue that the bias-only model should reflect informative statistics via subword frequency and design a class prior.
% Herein, RegLER takes advantage of predicting long-named and complex structure entities with bias-weighted scaling.
% Our experimental results also demonstrate the effectiveness and generalizability of the RegLER.
% To sum up, training a target model with three components can make a more robust model with a high capacity for generalization and memorization.
In this study, we introduce a debiasing method RegLER to regularize PLMs to predict well for various lengths of entities.
We evaluated our models on partitioned benchmark datasets containing unseen mention sets to improve practicality in real-world situations.
RegLER shows a significant improvement when predicting long-named entities while preserving performance on short-named entities.
While the debiasing framework is vulnerable to preserving memorization ability and in-domain performance, our method achieves competitive performance on all datasets.
In our observation, a combination of modeling a class prior and debiasing on subword frequency improves the generalization ability of synonym and concept mention sets.
We hope our method is meaningful to those who are working to improve the generalization ability of models.

\section{Ethical Considerations}
In this work, we focus on helping a wide range of existing Named Entity Recognition (NER) models to alleviate data bias and enhance generalization behavior.
There may exist many different biases, however, we attempt to mitigate the class imbalance problem inherent in the benchmark NER datasets.
As a result, our work promotes robustness with a high capacity for generalization and reduces concerns regarding manually annotating entities suffering from high class imbalance.
To save energy, we will share our code and dataset publicly, and also suggest hyperparameters for each experiment to prevent additional trials and reduce further carbon emissions.

\bibliography{anthology,acl}
\bibliographystyle{acl_natbib}

\appendix

\section{Dataset Statistics}
\label{app:dataset statistic}

Following the previous works~\cite{lin2020rigourous,kim2021your}, in Table~\ref{tab:partition dataset}, we partition the top three datasets that suggest the concept unique ID (CUI) in order to split the unseen mention set into \textsc{Syn} and \textsc{Con}.
Without the CUI, we partition the benchmark datasets into \textsc{Mem} and \textsc{Unseen} mention sets. 

\begin{table}[h]
\centering
\begin{adjustbox}{width=0.48\textwidth}
\begin{tabular}{l c c c c c c }
\toprule
\multicolumn{1}{c}{\multirow{2}{*}{Dataset}} & \multicolumn{3}{c}{Dev}              & \multicolumn{3}{c}{Test}              \\ \cmidrule{2-7} 
\multicolumn{1}{c}{} & \textsc{Mem}     & \textsc{Syn}          & \textsc{Con}          & \textsc{Mem}      & \textsc{Syn}          & \textsc{Con}          \\ \midrule
NCBI-disease             & 514     & 184          & 87           & 594      & 196          & 164          \\ 
BC5CDR-disease           & 2624    & 899          & 632          & 2802     & 882          & 631          \\ 
BC5CDR-chem              & 3432    & 460          & 1394         & 3293     & 515          & 1538         \\ \midrule
 & \textsc{Mem}     & \multicolumn{2}{c}{\textsc{Unseen}} & \textsc{Mem}      & \multicolumn{2}{c}{\textsc{Unseen}} \\ \midrule
BC4CHEMD                 & 18337   & \multicolumn{2}{c}{11082}  & 15974    & \multicolumn{2}{c}{9324}   \\ 
BC2GM                    & 1250    & \multicolumn{2}{c}{1810}   & 2625     & \multicolumn{2}{c}{3697}   \\ 
JNLPBA                   & 6356    & \multicolumn{2}{c}{2214}   & 3616     & \multicolumn{2}{c}{2614}   \\ 
LINNAEUS                 & 516     & \multicolumn{2}{c}{189}    & 976      & \multicolumn{2}{c}{448}    \\ 
Species-800              & 130     & \multicolumn{2}{c}{254}    & 270      & \multicolumn{2}{c}{494}    \\ 
CoNLL-2003               & 3963    & \multicolumn{2}{c}{1854}   & 2993     & \multicolumn{2}{c}{2500}   \\ \bottomrule
\end{tabular}
\end{adjustbox}
\caption{Partitioned benchmark datasets to evaluate generalization ability following the previous works \protect\cite{lin2020rigourous,kim2021your}}
\label{tab:partition dataset}
\end{table}

\section{Learned-Mixin and Learned-Mixin+H}
\label{app:learned-mixin}
Following the previous work~\cite{clark2019don}, we use an ensemble framework that combines a target model $f_\theta$ and a bias-only model $\frac{P(\textbf{Y}|\textbf{X})}{P(\textbf{Y})}$ as a statistic.
We also extend our ensemble methods to Learned-Mixin and Learned-Mixin+H without utilizing a bias product.
We compute a final probability as below:

\begin{equation}
    g(x_i) = softplus(w \cdot h_i),
\end{equation}

\begin{equation}
    \hat{p_{i}} = softmax(log(p_i) + g(x_i) \cdot log(b_i)),
\end{equation}

where $g$ is a learned function with a learned vector $w$ and the final hidden layer $h_i$ of the target model for the token $x_i$. 
Herein, a $softplus(x) = log(1+e^x)$ function is used to prevent the target model from reversing the bias~\cite{clark2019don}.

However, the Learned-Mixin method tends to ignore a bias probability $log(b_i)$.
Thus, \cite{clark2019don} also suggested the Learned-Mixin+H method which adds the entropy penalty to the objective function, preventing the target model from ignoring the bias probability $log(b_i)$.
An objective function with entropy penalty is computed as follows:

\begin{equation}
    R = w \cdot H(softmax(g(x_i) \cdot log(b_i)))
\end{equation}
\begin{equation}
    L = -\frac{1}{N}\sum_{i=1}^{N} y_{i} \cdot \hat{p_{i}} + R, \\
\end{equation}

where $H(z) = -\sum_kz_klog(z_k)$ is the entropy and $w$ is a hyperparameter (we select 0.2 in our experiments). 

A Learned-Mixin framework that prevents the target model from reversing the bias decreases generalizability while increasing memorization ability.
The experimental results show a similar trend to the bias-product framework thus we did not report this in our table.
Furthermore, the Learned-Mixin+H method shows unstable results because the two regularization functions (i.e., entropy penalty and temperature scaling) conflict with each other in a bias-only model.
In other words, the entropy penalty encourages the bias-only model to be non-uniform, but the temperature scaling forces the model to be uniform.
Owing to the low and unstable results, we removed these two methods from our results.

\begin{table*}[t]
\begin{adjustbox}{width=\textwidth}
\begin{tabular}{l c c c c c c | c c}
\toprule
 \multicolumn{1}{c}{\multirow{3}{*}{Method}} & \multicolumn{1}{c}{\multirow{3}{*}{\textsc{Subword}~}} &
\multicolumn{1}{c}{\multirow{3}{*}{\textsc{Class}~}} &
\multicolumn{1}{c}{\multirow{3}{*}{\textsc{Temp}~}} & \multicolumn{3}{c}{CoNLL 2003} & \multicolumn{2}{c}{WNUT 2017}      \\ \cmidrule{5-9} 
& & & & \textsc{Mem} (R) &  \textsc{Unseen} (R) &  Total (F1) & Total (F1) & Surface (F1) \\ \midrule
$ACE+document-context$ \cross & \xmark & \xmark & \xmark & - & - & \textbf{94.6} & - & - \\
$CL-KL$ \cross  & \xmark & \xmark & \xmark & - & - & 93.9 & - & \textbf{60.5} \\
$FLERT XLM-R$ \cross  & \xmark & \xmark & \xmark & - & - & 94.1 & - & - \\
$LUKE$ \cross & \xmark & \xmark & \xmark & - & - & 94.3 & - & - \\
\midrule
$BERT-base$ & \xmark & \xmark & \xmark & 94.7 & 88.0  & 91.0 & 45.1 & 43.1 \\
$BERT-base + {BP}$ & \xmark & \xmark & \xmark & 93.8 & 88.4  & 90.9 & 44.3 & 42.3 \\
$BERT-base$ + RegLER & \cmark & \cmark & \cmark & 95.2 & 89.1  & 91.5 & 47.8 & 46.6 \\
\midrule
$RoBERTa-base$ & \xmark & \xmark & \xmark & 94.1 & 88.7 & 90.3 & 52.4 & 50.1 \\
$RoBERTa-base + {BP}$ & \xmark & \xmark & \xmark & 93.5 & 88.7 & 90.8 & 52.5 & 50.8 \\
$RoBERTa-base$ + RegLER & \cmark & \cmark & \cmark & 94.4 & 89.1 & 90.8 & 52.8 & 51.2 \\
\midrule
$BERT-large$ & \xmark & \xmark & \xmark & 94.9 & 89.4 & 91.9 & 49.6 & 47.7 \\
$BERT-large + {BP}$ & \xmark & \xmark & \xmark & 94.7 & 89.6 & 91.9 & 49.9 & 48.1 \\
$BERT-large$ + RegLER & \cmark & \cmark & \cmark & 95.2 & 89.8 & 92.0 & 50.7 & 49.3 \\
\midrule
$RoBERTa-large$ & \xmark & \xmark & \xmark & 95.6 & 90.0 & 91.8 & 57.1 & 55.9 \\
$RoBERTa-large + {BP}$ & \xmark & \xmark & \xmark & 95.2 & 90.4 & 92.4 & 57.5 & 56.1 \\
$RoBERTa-large$ + RegLER & \cmark & \cmark & \cmark & \textbf{95.7} & \textbf{90.8} & 92.3 & \textbf{58.9} & 58.3 \\
\bottomrule
\end{tabular}
\end{adjustbox}
\caption{Performance of RegLER on the general domain NER datasets.}
\label{tab:maintable 2}
\end{table*}

\section{Hyperparameter and Training Details}
\label{app:hyperparameter}
We describe our hyperparameter and training details.
For all training models, we used at batch size of 32 and the Adam optimizer~\cite{kingma2014adam} with a learning rate in the range of \{5e-6, 1e-5, 3e-5\}.
We used a warmup rate in the range of \{0, 5000\} steps for all datasets and a training seed in the range of \{1, 2, 3, 4, 5\} to conduct five runs.
The length of the input sequence was set to 128 in the biomedical domain and 256 in the general domain dataset. 
If the sentence length is longer than 128 or 256, respectively, then the sentence was divided into multiple sentences.
A training epoch was selected in the range of \{20, 50\}.
To adopt temperature scaling~\cite{guo2017calibration} depending on the entity length, we used a lambda ($\lambda$) in the range of \{0.01, 0.1\}. 
For training and inference of all the experimental results, we use one graphics processing unit (NVIDIA Titan Xp).
Our code is based on the PyTorch implementation of \cite{lee2020biobert}.\footnote{https://github.com/dmis-lab/biobert-pytorch}
Most of the previous studies used a training dataset containing the training and development dataset in the BioNER task. 
In our experiments, we followed previous works to use the development dataset in our training process for the biomedical NER dataset.

\section{Experimental Results of the General Domain}
\label{app:general domain}

We used BERT~\cite{devlin2019bert} and RoBERTa~\cite{liu2019roberta} as our target models.
We enumerate our baselines as follows:
ACE$+$document-context~\cite{wang2020automated}, LUKE~\cite{yamada2020luke}, and FLERTXLM-R~\cite{schweter2020flert}.
For the evaluation, the WNUT-2017 dataset has its own evaluation metric that measures how good systems are at correctly recognizing a diverse range of entities, rather than simply measuring the very frequent surface forms.

On the CoNLL 2003 dataset, because RegLER aims to predict long-named entities, improvements in in-domain and generalization abilities (\textsc{Mem} and \textsc{Unseen}) are not effective.
However, RegLER shows a significant improvement on the WNUT 2017 dataset, which is composed of unseen mentions on the test dataset.
Compared to the fine-tuning language model (the first fow in each block), the absolute improvements of RegLER in the in-domain and surface form performance are approximately 1.5\% and 2.15\%, respectively.
In addition, the absolute improvements of RegLER for the in-domain and surface form performances are approximately 1.5\% and 2.0\%, respectively compared to \cite{kim2021your}. 
Based on these results, we believe that using RegLER shows little degradation in the surface form performance compared to the in-domain performance.
This means that our method learns various mention forms, rather than focusing on a specific mention form.

\begin{table*}[t]
\begin{adjustbox}{width=\textwidth}
\begin{tabular}{ c l l l l l l l l }
\toprule
\multicolumn{9}{c}{BC2GM} \\ \midrule
\multicolumn{1}{c}{$\lambda$} & \multicolumn{1}{c}{\begin{tabular}[c]{@{}c@{}}E(L) = 1\\ (P/R/F)\end{tabular}} & \multicolumn{1}{c}{\begin{tabular}[c]{@{}c@{}}E(L) = 2\\ (P/R/F)\end{tabular}} & \multicolumn{1}{c}{\begin{tabular}[c]{@{}c@{}}E(L) = 3\\ (P/R/F)\end{tabular}} & \multicolumn{1}{c}{\begin{tabular}[c]{@{}c@{}}E(L) = 4\\ (P/R/F)\end{tabular}} & \multicolumn{1}{c}{\begin{tabular}[c]{@{}c@{}}E(L) = 5\\ (P/R/F)\end{tabular}} & \multicolumn{1}{c}{\begin{tabular}[c]{@{}c@{}}E(L) = 6\\ (P/R/F)\end{tabular}} & \multicolumn{1}{c}{\begin{tabular}[c]{@{}c@{}}E(L) = 7\\ (P/R/F)\end{tabular}} & \multicolumn{1}{c}{\begin{tabular}[c]{@{}c@{}}E(L) $\geqq$ 8\\ (P/R/F)\end{tabular}} \\ \midrule
1.09                                         & \multicolumn{1}{c}{87.4 / 85.6 / 86.5}                                         & \multicolumn{1}{c}{85.3 / 83.4 / 84.1}                                         & \multicolumn{1}{c}{86.2 / 85.9 / 86.0}                                         & \multicolumn{1}{c}{83.6 / 83.5 / 83.5}                                         & \multicolumn{1}{c}{81.4 / 81.6 / 81.5}                                         & \multicolumn{1}{c}{82.2 / 82.1 / 82.2}                                         & \multicolumn{1}{c}{80.4 / 78.8 / 79.6}                                         & \multicolumn{1}{c}{69.6 / 74.1 / 71.8}                                         \\
1.08 & 87.1 / 85.3 / 86.2 & 85.1 / 83.3 / 84.2 & 86.0 / 85.5 / 85.8 & 83.6 / 83.3 / 83.4 & 81.7 / 81.5 / 81.6 & 81.5 / 81.8 / 81.6 & 80.5 / 79.0 / 79.7 & 70.1 / 73.8 / 72.0 \\ 
1.07 & 87.6 / 85.9 / 86.7 & 85.4 / 83.7 / 84.6 & 86.1 / 85.7 / 85.9 & 83.9 / 83.7 / 83.8 & 82.0 / 81.7 / 81.8 & 81.2 / 81.6 / 81.4 & 79.7 / 78.8 / 79.3 & 70.7 / 73.8 / 72.2 \\ 
1.06                                         & 87.5 / 86.3 / 86.9 & 85.4 / 83.9 / 84.6 & 85.7 / 85.7 / 85.7 & 83.4 / 83.6 / 83.5 & 82.1 / 82.5 / 82.3 & 81.6 / 82.1 / 81.9 & 80.1 / 80.4 / 80.3 & 70.1 / 73.6 / 71.8 \\ 
1.05                                         & 87.6 / 86.3 / 86.9 & 85.5 / 83.9 / 84.7 & 86.2 / 86.0 / 86.1 & 83.6 / 83.5 / 83.6 & 81.8 / 81.9 / 81.8 & 81.2 / 82.2 / 81.7 & 80.2 / 79.7 / 79.8 & 69.4 / 73.5 / 71.4 \\ 
1.04                                         & 87.6 / 86.1 / 86.8 & 85.6 / 84.0 / 84.8 & 86.1 / 85.7 / 85.9 & 83.7 / 83.5 / 83.6 & 81.8 / 81.7 / 81.7 & 82.3 / 83.0 / 82.6 & 80.3 / 79.8 / 80.0 & 71.6 / 74.7 / 73.1 \\ 
1.03                                         & 87.2 / 85.8 / 86.5 & 85.7 / 83.9 / 84.8 & 81.2 / 85.9 / 86.0 & 84.1 / 83.8 / 84.0 & 81.9 / 81.8 / 81.9 & 82.0 / 82.5 / 82.3 & 80.9 / 80.8 / 80.9 & 71.5 / 74.5 / 72.9 \\ 
1.02                                         & 87.5 / 86.0 / 86.7 & 85.3 / 83.7 / 84.5 & 86.1 / 85.7 / 85.9 & 83.5 / 83.4 / 83.5 & 82.1 / 81.9 / 82.0 & 81.5 / 82.0 / 81.8 & 79.4 / 78.7 / 79.0 & 70.2 / 73.3 / 71.7 \\ 
1.01                                         & 87.0 / 85.2 / 86.1 & 85.0 / 83.3 / 84.1 & 85.8 / 85.5 / 85.7 & 83.2 / 83.4 / 83.3 & 82.2 / 82.2 / 82.2 & 80.5 / 81.6 / 81.0 & 79.2 / 79.3 / 79.3 & 69.4 / 72.8 / 71.1 \\ \bottomrule
\end{tabular}
\end{adjustbox}
\caption{Ablation study of using various $\lambda$ values. E(L) signifies the length of the entity.}
\label{tab:app length ablation}
\end{table*}

\section{Experimental Results of Other Biomedical Datasets}
\label{app:experimental results}

Table~\ref{tab:maintable 3} shows our results for the rest of the biomedical NER datasets.
The main trend is similar in that our approach enhances the recall of \textsc{Mem} and \textsc{Unseen} while preserving (or sometimes improving) the in-domain performance.
Our RegLER method shows competitive results on the BC4CHEMD, BC2GM, and BC5CDR-disease datasets.
We conducted five runs with different seeds for all the experiments. The t-test indicated that p $<$ 0.05.

\begin{table*}[t]
\centering
\begin{adjustbox}{width=0.75\textwidth}
\begin{tabular}{ l l c c c c c c }
\toprule
\multicolumn{1}{c}{\multirow{2}{*}{Dataset}} & \multicolumn{1}{c}{\multirow{2}{*}{Method}} & \multicolumn{1}{c}{\multirow{2}{*}{\textsc{Subword}~}} & \multicolumn{1}{c}{\multirow{2}{*}{\textsc{Class}~}} &
\multicolumn{1}{c}{\multirow{2}{*}{\textsc{Temp}~}} & \multicolumn{3}{c}{Eval Metric}       \\ \cmidrule{6-8} 
\multicolumn{1}{c}{} & & & & & 
\textsc{Mem} (R) & 
\textsc{Unseen} (R) & 
Total (F1) \\ \midrule
\multirow{21}{*}{BC4CHEMD}                   
& $SciFive-Base$ & \xmark & \xmark & \xmark & - & - & - \\
& $SparkNLP$ & \xmark & \xmark & \xmark & - & - & \textbf{93.7} \\
& $KeBioLM$ & \xmark & \xmark & \xmark & - & - & -  \\
& $CL - KL$ & \xmark & \xmark & \xmark & - & - & -  \\
& $BioFLAIR$ & \xmark & \xmark & \xmark & - & - & - \\
\cmidrule{2-8}
& $BioBERT$ & \xmark & \xmark & \xmark & 95.6 & 82.6 & 91.1 \\
& $BioBERT + {BP}$ & \xmark & \xmark & \xmark & 94.9 & 83.3 & 91.0 \\
& $BioBERT$ + RegLER & \cmark & \cmark & \cmark & 95.8 & 83.9 & 91.4 \\
\cmidrule{2-8}
& $SciBERT$ & \xmark & \xmark & \xmark & 94.2 & 79.2 & 89.4 \\
& $SciBERT + {BP}$ & \xmark & \xmark & \xmark & 94.2 & 80.5 & 89.4 \\
& $SciBERT$ + RegLER & \cmark & \cmark & \cmark & 94.2 & 81.1 & 89.4 \\
\cmidrule{2-8}
& $PubmedBERT$ & \xmark & \xmark & \xmark & 95.3 & 82.8 & 91.2 \\
& $PubmedBERT + {BP}$ & \xmark & \xmark & \xmark & 95.3 & 83.7 & 91.2 \\
& $PubmedBERT$ + RegLER & \cmark & \cmark & \cmark & 95.5 & 83.5 & 91.4 \\
\cmidrule{2-8}
& $BioLM-base$ & \xmark & \xmark & \xmark & 95.4 & 86.5 & 91.4 \\
& $BioLM-base + {BP}$ & \xmark & \xmark & \xmark & 95.1 & 86.6 & 91.1 \\
& $BioLM-base$ + RegLER & \cmark & \cmark & \cmark & 96.0 & 87.1 & 92.5 \\
\cmidrule{2-8}
& $BioLM-large$ & \xmark & \xmark & \xmark & \textbf{96.3} & 88.2 & 93.2 \\
& $BioLM-large + {BP}$ & \xmark & \xmark & \xmark & \textbf{96.3} & 88.5 & 93.4 \\
& $BioLM-large$ + RegLER & \cmark & \cmark & \cmark & 96.2 & \textbf{88.9} & 93.6 \\
\midrule
\multirow{21}{*}{BC2GM}
& $SciFive-Base$ & \xmark & \xmark & \xmark & - & - & - \\
& $SparkNLP$ & \xmark & \xmark & \xmark & - & - & -  \\
& $KeBioLM$ & \xmark & \xmark & \xmark & - & - & \textbf{85.1}  \\
& $CL - KL$ & \xmark & \xmark & \xmark & - & - & -  \\
& $BioFLAIR$ & \xmark & \xmark & \xmark & - & - & - \\
\cmidrule{2-8}
& $BioBERT$ & \xmark & \xmark & \xmark & 89.8 & 80.2 & 83.5 \\
& $BioBERT + {BP}$ & \xmark & \xmark & \xmark & 89.2 & 80.2 &  83.4 \\
& $BioBERT$ + RegLER & \cmark & \cmark & \cmark & 89.7 & 80.6 & 83.7 \\
\cmidrule{2-8}
& $SciBERT$ & \xmark & \xmark & \xmark & 88.6 & 77.5 & 82.1 \\
& $SciBERT + {BP}$ & \xmark & \xmark & \xmark & 88.1 & 78.4 & 82.0 \\
& $SciBERT$ + RegLER & \cmark & \cmark & \cmark & 88.5 & 78.9 & 82.4 \\
\cmidrule{2-8}
& $PubmedBERT$ & \xmark & \xmark & \xmark & 89.9 & 81.1 & 84.1 \\
& $PubmedBERT + {BP}$ & \xmark & \xmark & \xmark & 89.1 & 81.1 & 83.8 \\
& $PubmedBERT$ + RegLER & \cmark & \cmark & \cmark & 90.7 & 81.8 & 84.4 \\
\cmidrule{2-8}
& $BioLM-base$ & \xmark & \xmark & \xmark & 90.5 & 81.8 & 83.5 \\
& $BioLM-base + {BP}$ & \xmark & \xmark & \xmark & 90.2 & 82.1 & 82.9 \\
& $BioLM-base$ + RegLER & \cmark & \cmark & \cmark & \textbf{90.8} & \textbf{82.8} & 84.7 \\
\cmidrule{2-8}
& $BioLM-large$ & \xmark & \xmark & \xmark & 90.7 & 82.1 & 84.6 \\
& $BioLM-large + {BP}$ & \xmark & \xmark & \xmark & \textbf{90.8} & 82.1 & \textbf{85.1} \\
& $BioLM-large$ + RegLER & \cmark & \cmark & \cmark & \textbf{90.8} & 82.7 & \textbf{85.1} \\
\midrule
\multirow{21}{*}{JNLPBA}
& $SciFive-Base$ & \xmark & \xmark & \xmark & - & - & - \\
& $SparkNLP$ & \xmark & \xmark & \xmark & - & - & 81.3  \\
& $KeBioLM$ & \xmark & \xmark & \xmark & - & - & \textbf{82.0}  \\
& $CL - KL$ & \xmark & \xmark & \xmark & - & - & -  \\
& $BioFLAIR$ & \xmark & \xmark & \xmark & - & - & 77.0 \\
\cmidrule{2-8}
& $BioBERT$ & \xmark & \xmark & \xmark & 89.7 & 73.7 &  76.2 \\
& $BioBERT + {BP}$ & \xmark & \xmark & \xmark & 89.4 & 74.1 &  76.4 \\
& $BioBERT$ + RegLER & \cmark & \cmark & \cmark & 89.8 & 74.3 & 76.7 \\
\cmidrule{2-8}
& $SciBERT$ & \xmark & \xmark & \xmark & 89.7 & 73.5 & 76.1 \\
& $SciBERT + {BP}$ & \xmark & \xmark & \xmark & 89.3 & 74.2 & 76.4 \\
& $SciBERT$ + RegLER & \cmark & \cmark & \cmark & 89.7 & 74.6 &  76.4 \\
\cmidrule{2-8}
& $PubmedBERT$ & \xmark & \xmark & \xmark & 90.2 & 74.7 & 77.3 \\
& $PubmedBERT + {BP}$ & \xmark & \xmark & \xmark & 89.3 & 75.2 & 77.1 \\
& $PubmedBERT$ + RegLER & \cmark & \cmark & \cmark & 89.8 & 75.7 & 77.2 \\
\cmidrule{2-8}
& $BioLM-base$ & \xmark & \xmark & \xmark & \textbf{90.3} & 75.7 & 76.8 \\
& $BioLM-base + {BP}$ & \xmark & \xmark & \xmark & 89.6 & 76.0 & 76.2 \\
& $BioLM-base$ + RegLER & \cmark & \cmark & \cmark & 90.0 & \textbf{76.5} & 77.3 \\
\cmidrule{2-8}
& $BioLM-large$ & \xmark & \xmark & \xmark & 89.2 & 73.5 & 76.4 \\
& $BioLM-large + {BP}$ & \xmark & \xmark & \xmark & 89.5 & 74.3 & 76.2 \\
& $BioLM-large$ + RegLER & \cmark & \cmark & \cmark & 90.2 & 75.7 & 76.8 \\
\bottomrule
\end{tabular}
\end{adjustbox}
\caption{Performance of the debiasing method RegLER on the biomedical domain NER datasets. Each dataset is partitioned into a memorization (\textsc{Mem}) and \textsc{Unseen} mention set. Best performances are shown in bold.}
\label{tab:maintable 3}
\end{table*}

\begin{table*}[t]
\centering
\begin{adjustbox}{width=0.75\textwidth}
\begin{tabular}{ l l c c c c c c c }
\toprule
\multicolumn{1}{c}{\multirow{2}{*}{Dataset}} & \multicolumn{1}{c}{\multirow{2}{*}{Method}} & \multicolumn{1}{c}{\multirow{2}{*}{\textsc{Subword}~}} & \multicolumn{1}{c}{\multirow{2}{*}{\textsc{Class}~}} &
\multicolumn{1}{c}{\multirow{2}{*}{\textsc{Temp}~}} & \multicolumn{4}{c}{Eval Metric}       \\ \cmidrule{6-9} 
\multicolumn{1}{c}{} & & & & & 
\textsc{Mem} (R) & 
\textsc{Unseen} (R) & 
\multicolumn{2}{c}{Total (F1)} \\ \midrule
\multirow{21}{*}{S800}
& $SciFive-Base$ & \xmark & \xmark & \xmark & - & - & \multicolumn{2}{c}{-} \\
& $SparkNLP$ & \xmark & \xmark & \xmark & - & - & \multicolumn{2}{c}{80.9}  \\
& $KeBioLM$ & \xmark & \xmark & \xmark & - & - & \multicolumn{2}{c}{-} \\
& $CL - KL$ & \xmark & \xmark & \xmark & - & - & \multicolumn{2}{c}{-} \\
& $BioFLAIR$ & \xmark & \xmark & \xmark & - & - & \multicolumn{2}{c}{\textbf{82.4}} \\
\cmidrule{2-9}
& $BioBERT$ & \xmark & \xmark & \xmark & 93.6 & 70.6 & \multicolumn{2}{c}{74.7} \\
& $BioBERT + {BP}$ & \xmark & \xmark & \xmark & 93.8 & 71.4 & \multicolumn{2}{c}{74.4} \\
& $BioBERT$ + RegLER & \cmark & \cmark & \cmark & \textbf{94.1} & 72.2 & \multicolumn{2}{c}{75.3} \\
\cmidrule{2-9}
& $SciBERT$ & \xmark & \xmark & \xmark & 89.5 & 66.5 & \multicolumn{2}{c}{72.0} \\
& $SciBERT + {BP}$ & \xmark & \xmark & \xmark & 89.3 & 65.4 & \multicolumn{2}{c}{71.7} \\
& $SciBERT$ + RegLER & \cmark & \cmark & \cmark & 90.7 & 68.2 & \multicolumn{2}{c}{70.9} \\
\cmidrule{2-9}
& $PubmedBERT$ & \xmark & \xmark & \xmark & 90.5 & 72.2 & \multicolumn{2}{c}{74.2} \\
& $PubmedBERT + {BP}$ & \xmark & \xmark & \xmark & 89.3 & 72.2 & \multicolumn{2}{c}{73.2} \\
& $PubmedBERT$ + RegLER & \cmark & \cmark & \cmark & 91.1 & 72.4 & \multicolumn{2}{c}{73.6} \\
\cmidrule{2-9}
& $BioLM-base$ & \xmark & \xmark & \xmark & 92.4 & 70.9 &  \multicolumn{2}{c}{73.2} \\
& $BioLM-base + {BP}$ & \xmark & \xmark & \xmark & 91.9 & 72.1 & \multicolumn{2}{c}{72.8} \\
& $BioLM-base$ + RegLER & \cmark & \cmark & \cmark & 92.2 & 71.7 & \multicolumn{2}{c}{73.9} \\
\cmidrule{2-9}
& $BioLM-large$ & \xmark & \xmark & \xmark & 90.2 & 73.3 & \multicolumn{2}{c}{73.3} \\
& $BioLM-large + {BP}$ & \xmark & \xmark & \xmark & 91.5 & \textbf{74.3} & \multicolumn{2}{c}{73.7} \\
& $BioLM-large$ + RegLER & \cmark & \cmark & \cmark & 92.2 & 73.7 & \multicolumn{2}{c}{74.5} \\ \midrule \midrule
\multicolumn{1}{c}{\multirow{2}{*}{Dataset}} & \multicolumn{1}{c}{\multirow{2}{*}{Method}} & \multicolumn{1}{c}{\multirow{2}{*}{\textsc{Subword}~}} & \multicolumn{1}{c}{\multirow{2}{*}{\textsc{Class}~}} &
\multicolumn{1}{c}{\multirow{2}{*}{\textsc{Temp}~}} & \multicolumn{4}{c}{Eval Metric}       \\ \cmidrule{6-9} 
\multicolumn{1}{c}{} & & & & & 
\textsc{Mem} (R) & 
\textsc{Syn} (R) & 
\textsc{Con} (R) & 
Total (F1) \\
\midrule
\multirow{21}{*}{\begin{tabular}[c]{@{}c@{}}BC5CDR-\\ disease\end{tabular}}
& $SciFive-Base$ & \xmark & \xmark & \xmark & - & - & - & \textbf{87.2} \\
& $SparkNLP$ & \xmark & \xmark & \xmark & - & - & - & -  \\
& $KeBioLM$ & \xmark & \xmark & \xmark & - & - & - & 86.1 \\
& $CL - KL$ & \xmark & \xmark & \xmark & - & - & - & -  \\
& $BioFLAIR$ & \xmark & \xmark & \xmark & - & - & - & 85.3 \\
\cmidrule{2-9}
& $BioBERT$ & \xmark & \xmark & \xmark & 91.3 & 76.7 & 73.4 & 82.5 \\
& $BioBERT + {BP}$ & \xmark & \xmark & \xmark & 90.9 & 77.1 & 74.3 & 82.5 \\
& $BioBERT$ + RegLER & \cmark & \cmark & \cmark & 92.0 & 78.5 & 75.1 & 83.3 \\
\cmidrule{2-9}
& $SciBERT$ & \xmark & \xmark & \xmark & 93.3 & 71.9 & 73.3 & 84.2  \\
& $SciBERT + {BP}$ & \xmark & \xmark & \xmark & 93.2 & 70.8 & 73.3 & 84.1  \\
& $SciBERT$ + RegLER & \cmark & \cmark & \cmark & 93.1 & 72.2 & 74.2 & 84.1 \\
\cmidrule{2-9}
& $PubmedBERT$ & \xmark & \xmark & \xmark & 93.2 & 76.5 & 74.5 & 85.4 \\
& $PubmedBERT + {BP}$ & \xmark & \xmark & \xmark & 93.1 & 78.6 & 75.7 & 85.3 \\
& $PubmedBERT$ + RegLER & \cmark & \cmark & \cmark & 93.4 & 77.5 & 75.3 & 85.5 \\
\cmidrule{2-9}
& $BioLM-base$ & \xmark & \xmark & \xmark & 93.5 & 79.2 & 75.6 & 85.4 \\
& $BioLM-base + {BP}$ & \xmark & \xmark & \xmark & 92.5 & 79.8 & 76.7 & 83.9 \\
& $BioLM-base$ + RegLER & \cmark & \cmark & \cmark & 93.5 & 80.8 & 77.5 & 86.2 \\
\cmidrule{2-9}
& $BioLM-large$ & \xmark & \xmark & \xmark & 93.5 & 81.5 & 77.4 & 86.2 \\
& $BioLM-large + {BP}$ & \xmark & \xmark & \xmark & 93.0 & 82.3 & 77.4 & 86.4 \\
& $BioLM-large$ + RegLER & \cmark & \cmark & \cmark & \textbf{93.7} & \textbf{83.2} & \textbf{78.5} & 86.9 \\
\bottomrule
\end{tabular}
\end{adjustbox}
\caption{Performance of the debiasing method RegLER on the biomedical domain NER datasets. Each dataset is partitioned into a memorization (\textsc{Mem}) and \textsc{Unseen} mention set. Best performances are shown in bold.}
\label{tab:maintable 4}
\end{table*}

\section{Experimental Results of Various Lambda}
\label{app:lambda results}

As stated, Table~\ref{tab:app length ablation} shows that in a mention set with a small entity length (i.e., E(L) is low), there is no significant difference in performance even if $\lambda$ changes.
However, we find that a mention set of long-named entities (i.e., E(L) $\geqq$ 8) exhibits at high-performance gap between low $\lambda$ and high $\lambda$ values.
Because it is difficult to find an optimal $\lambda$ for each mention set, finding it through a validation setting or set a $\lambda$ as a learnable parameter appears to be a good approach.

\end{document}